\documentclass[lettersize,journal]{IEEEtran}
\usepackage{amsmath,amssymb}
\usepackage{algorithmic}
\usepackage{array}
\usepackage{footnote}
\usepackage{booktabs,makecell,tabularx}
\usepackage{multirow}
\usepackage{subcaption}
\usepackage{textcomp}
\usepackage{stfloats}
\usepackage{url}
\usepackage{verbatim}
\usepackage{graphicx}
\usepackage[mathscr]{euscript}

\def\BibTeX{{\rm B\kern-.05em{\sc i\kern-.025em b}\kern-.08em
    T\kern-.1667em\lower.7ex\hbox{E}\kern-.125emX}}
\usepackage{balance}

\begin{document}
\title{VTAE: Variational Transformer Autoencoder with Manifolds Learning}
\author{Pourya~Shamsolmoali,~\IEEEmembership{Member,~IEEE,}
        Masoumeh~Zareapoor, Huiyu~Zhou,
        Dacheng~Tao,~\IEEEmembership{Fellow,~IEEE},
        and~Xuelong~Li,~\IEEEmembership{Fellow,~IEEE}
\thanks{Manuscript submitted June, 2022. }

\thanks{ P.~Shamsolmoali is with the School of Communication and Electrical Engineering, East China Normal University, Shanghai, China (pshams55@gmail.com).}
\thanks{M.~Zareapoor is with the School of Electronic Information and Electrical Engineering, Shanghai Jiao Tong University, (mzarea222@gmail.com).}
\thanks{H.~Zhou is with the School of Computing and Mathematical Sciences, University of Leicester, Leicester LE1 7RH, UK (hz143@leicester.ac.uk).}
\thanks{D.~Tao is with the School of Computer Science, Faculty of Engineering, The University of Sydney, Darlington, NSW, Australia (dacheng.tao@sydney.edu.au).}
\thanks{X.~Li is with the School of Artificial Intelligence, OPtics and ElectroNics (iOPEN), Northwestern Polytechnical University, and also with the Key Laboratory of Intelligent Interaction and Applications (Northwestern Polytechnical University), Ministry of Industry and Information Technology, Xi'an, China (li@nwpu.edu.cn).}}

\markboth{IEEE TRANSACTIONS ON IMAGE PROCESSING}%
{How to Use the IEEEtran \LaTeX \ Templates}

\maketitle

\begin{abstract}
Deep generative models have demonstrated successful
applications in learning non-linear data distributions through
a number of latent variables and these models use a non-linear
function (generator) to map latent samples into the data space.
On the other hand, the non-linearity of the generator implies
that the latent space shows an unsatisfactory projection of the
data space, which results in poor representation learning. This
weak projection, however, can be addressed by a Riemannian
metric, and we show that geodesics computation and accurate
interpolations between data samples on the Riemannian manifold
can substantially improve the performance of deep generative
models. In this paper, a Variational spatial-Transformer AutoEncoder
(VTAE) is proposed to minimize geodesics on a Riemannian
manifold and improve representation learning. In particular, we
carefully design the variational autoencoder with an encoded
spatial-Transformer to explicitly expand the latent variable model
to data on a Riemannian manifold, and obtain global context
modelling. Moreover, to have smooth and plausible interpolations
while traversing between two different objects’ latent
representations, we propose a geodesic interpolation network
different from the existing models that use linear interpolation
with inferior performance. Experiments on benchmarks show
that our proposed model can improve predictive accuracy and
versatility over a range of computer vision tasks, including image
interpolations, and reconstructions.
\end{abstract}


\section{Introduction}
\IEEEPARstart {G}{enerative} models are effective machine learning algorithms that are used to generate realistic synthetic data. The choice of data representation has a significant impact on generative models \cite{connor2021variational, yi2020bsd}. The performance of generative models is derived from the representation of data in more refined structures that are matched to those of the training data. Many generative models are designed to do a single task. For example, Generative Adversarial Networks (GANs) \cite{goodfellow2014generative} generate high-quality samples but do not allow for a simple assessment of data probability. They often fail to cover the whole data space with their samples. Variational autoencoders (VAEs) \cite{kingma2013auto} are a type of probabilistic models that generate a lower-dimensional encoded representation of features from which new data samples are generated. They are trained to maximize a lower bound on the data likelihood (ELBO). VAEs have expressive designs and need fine-tuning extra parameters, architectural adjustments, or unique training processes to promote better representation learning to seek acceptable sample quality \cite{alemi2018fixing}. \\
\vspace{-7pt}

 Representation learning, on the other hand, has an impact on more than just models performance. Transferable representations are useful for learning new tasks \cite{mikolov2013distributed}, representations are used to understand machine learning models, and usefully structured (disentangled) representations can also be used to manipulate models (e.g. semi-supervised learning \cite{kingma2014semi}). As a result, having interpretable data representations in a model is generally useful, in the sense that the information included in the representation is simply understood and the representation can be utilized to affect the model's output (For example, to produce data of a certain class or with a specific feature). Unfortunately, there is typically a trade-off between a model's best performance and its ability to be disentangled or controlled. To address this, some practitioners have proposed adjusting the objective functions of their models by introducing new hyperparameters, internal optimization loops, and controllers \cite{higgins2016beta, chen2018isolating, shao2022rethinking}, while others have suggested modifying the related generative model \cite{mansbridge2018improving} or developing a latent encoding via probing subspaces \cite{yang2021learning}. Additionally, efforts have also been made to incorporate data symmetries directly into a neural network architecture in order to induce the learning of latent variables that transforms significantly under certain symmetries \cite{sabour2017dynamic}. The complexity and limited success of these methods illustrate the critical nature and difficulty of learning meaningful representations in deep generative models.\\
\vspace{-7pt}

In non-linear latent variable models, the latent variables are often unidentified because the latent space is not invariant to reparametrizations which resulted in poor projection of the input data space. This identifiability problem is alleviated by enforcing a Riemannian metric on a smooth (Riemannian) manifold which considerably improves model fit, interpretability, and enhance learning \cite{chen2020learning}. Indeed, the Riemannian metric provides a more meaningful representation of the latent space and enhances probability distributions in the latent space.
Locally, points in the vicinity of a given latent vector can be sampled and decoded to analyse a region of space. On the other hand, global approaches, such as latent directions analysis, and interpolations, are intended to capture long distance relationships between points in the space.
The advantage of interpolation is the interpretability that comes with one-dimensional curves rather than high-dimensional Euclidean space. For instance, if a meaningful representation has been discovered, one would predict that the latent space represents the underlying structure of the dataset. In this case, decoded interpolation is used to support network generalisation capability and non-linear interpolations can eliminate the distribution mismatch between latent spaces, and provide a progressive transformation \cite{lesniak2018distribution}.\\
\vspace{-7pt}

\begin{figure}[t]
\includegraphics[width=0.88\columnwidth]{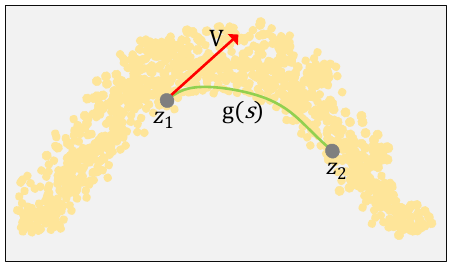} 
\centering
\caption{A data manifold comprising a geodesic between two points (green curve) and its tangent vector in a latent space.}   
\label{fig:1}
\end{figure}

In this paper, VTAE is proposed to obtain the geometric structure of a data manifold, estimate geodesic distance between data samples using Riemannian metric and construct a semantic feature space with transformation-specific interpretability to enhance the capacity of our model.
In practice, we design a new variation of VAE with an encoded Transformer to use the model's inference features for excellent representation learning by manipulating and learning the latent representation vectors. To extract a perspective invariant appearance, we first encode and decode the perspective characteristics and this is then encoded into the Transformer latent space, in which the inputs are encoded.
Moreover, to have a smooth manifold, the Euclidean space and learning a Riemannian metric are considered to directly measure distances between the data while preserving the structure of the data. An example is shown in Fig. \ref{fig:1}.
To do this, first, a geodesic learning approach is introduced to minimize geodesics curve between samples on a data manifold. 
Second, based on Riemannian geometry of data manifold \cite{shao2018riemannian}, we introduce a geodesic distance loss to compel the model to generate auxiliary samples alongside the geodesic curves. These auxiliary samples are generated in the domain of the transformed data samples, thus their spatial locations are incorporated.\\
 \vspace{-7pt}

Our model does not need any changes to the training process or new penalty terms in the objective function. It uses two latent variables, the first one for the invariant representation and the second for the smooth equivariant representation. For the equivariant representation, a Transformer layer is used to enhance disentanglement. Moreover, different from the current models \cite{higgins2016beta, chen2018isolating, jaderberg2015spatial, jiang2019linearized} that use linear or bilinear interpolation which are local, we propose a geodesic interpolation network to have smooth and plausible interpolations in the latent space. With these revolutionary changes, our model can reliably estimate complex latent distributions, which improves representation quality.
These contributions can be considered as the primary novel aspects of our approach. The main contributions of our work are twofold.
\begin{itemize}
\item{In order to learn the manifold structure from the data and use it to construct a right prior in the latent space, we propose a new variational autoencoder with an encoded Transformer to capture the global context information. To obtain the geometric information of data, the encoder extracts perspective invariant and equivariant features and encodes them into the Transformer's latent space, in which data with similar visual appearance are encoded identically. This results in an inductive bias which disentangled representations of visual data.}
\item{We develop a model to extend the latent variable model to data on Riemannian manifolds and compute the geodesic on Riemannian manifolds. We propose three loss functions for the interpolation network to construct the shortest path between two sample points, which provides semantically meaningful interpolations and leads to better image reconstruction quality.}
\end{itemize}

\noindent The remainder of this paper is structured as follows. Section II reports the related work. Section III describes the preliminaries. Section IV and V present the proposed framework that contains the geodesic manifold learning and a parametrized Transformer layer respectively that are encoded in our proposed model. The experimental results on six datasets are reported in Section VI to validate the performance of the proposed model. Finally Section VII concludes this paper.

\section{Related Work}
\noindent Given the wide generative models' literature, we focus only on the most recent and closely related studies.\\
\vspace{-7pt}

\noindent{\bf VAEs and Variants.}
VAE is a kind of generative model that creates new sample data from previous data. In order to understand the training data and their underlying distribution, VAEs map between latent variables. The vectors of latent variables are then used to reconstruct new sample data that is similar to the real data.
Conditional VAE (CVAE) \cite{sohn2015learning} is a modification of VAE to generate the images conditioned on the given attributes. CVAE has been widely used in text generation, especially for dialog generation \cite{ramchandran2022learning}. HyperVAE \cite{nguyen2021variational} is a variational inference architecture to encode the parameter distributions into a low dimensional Gaussian distribution. HyperVAE works at the class level, whereas the VAEs work at different design levels. The above models are computationally expensive. To alleviate this inefficiency, the convolution neural network (CNN) encoder of VAEs is replaced with a Gaussian process encoder in \cite{butepage2021gaussian} to provide latent uncertainty estimations for in-distribution and out-of-distribution data. Probabilistic autoencoder (PAE) \cite{bohm2020probabilistic} is a generative model from a lower dimensional latent space that can produce high quality samples but has a diminishing influence on the data distribution. \\
\vspace{-8pt}

Flat manifold VAE (FMVAE) \cite{chen2020learning} is an extension of VAE that enables the learning of flat latent manifolds using the Euclidean metric as a measure to compute the similarity between data samples. In \cite{ghosh2019variational}, a deterministic variant of VAEs (RAEs) is proposed that scales better, is considerably simpler to optimize, and generates better qualitative results than VAEs. In this model, regularization schemes are used instead of the noisy injection moulding to better learn latent space. \\
\vspace{-7pt}

$\beta$-VAE \cite{higgins2016beta} is another variation of the conventional VAE that is often used as an unsupervised approach for learning a disentangled representation of the generative components. Based on \cite{connor2021variational}, a disentangled representation is the one in which individual latent units are sensitive to changes in a generative component while remaining invariant to the changes of other variables. In comparison with VAEs, $\beta$-VAE incorporates one additional hyperparameter $\beta$, which serves as a weight for KL-divergence. 
To increase the capacity of $\beta$-VAE, AnnealedVAE \cite{burgess2018understanding} was proposed in which the encoder relies on progressive learning of the latent code during training.
Factor-VAE \cite{kim2018disentangling} is another technique that disentangles by promoting the factorial and dimension-independent distribution of representations. Factor-VAE outperforms $\beta$-VAE because it provides more optimal balance between disentanglement and reconstruction quality. To enhance Factor-VAE performance, in \cite{zhu2021commutative}, the authors proposed to encode data variations in groups to adaptively preserve the features of data variations.
The disadvantage of this approach is that reconstruction quality must be sacrificed to achieve better disentangling.
 In \cite{chen2018isolating}, $\beta$-TCVAE is proposed as a substitute for $\beta$-VAE that requires no additional hyperparameters. In $\beta$-TCVAE, the ELBO is decomposed to reveal the existence of a term measuring and the total correlation between latent variables. 
In these approaches, the compact prior is often characterized as a normal distribution. A normal prior, on the other hand, causes the approximate posterior to be over-regularized, leading to a less effective learned latent representation of the input. \\
\vspace{-7pt}

\noindent{\bf Manifold learning.}
The goal of manifold learning approaches is to infer latent representations that reflect the inherent geometric structure of data. Such methods are mainly used for data visualization and dimensionality reduction and recently are used to improve the performance of generative models \cite{deng2022gram, ni2022manifold, zhang2021geodesic, arvanitidis2017latent}. 
Traditional non-linear manifold learning methods like local tangent space alignment \cite{yang2005better} and linear embedding \cite{roweis2000nonlinear} maintain local structures and extract the features of data manifolds.
In \cite{kalatzis2020variational}, the Riemannian Brownian motion model was constructed over the manifold by using VAE (R-VAE). The Riemannian structure overcomes the identifiability problem by providing a meaningful representation which is not affected by reparametrizations. However, Riemannian manifold-based models cannot accurately find the shortest path between two samples that are far away from each other.
Generally, Riemannian geometry methods use Ordinary Differential Equation (ODE) to find the shortest interpolation curve in the latent space. ODE is computationally expensive and cannot find the shortest interpolation curve more consistently than straight-length methods.
The interpolation among data samples is widely used to attain a smooth transformation from one sample to another. 
In \cite{agustsson2017optimal}, the authors proposed to adopt distribution matching map transport to have a smooth interpolation and retain the intrinsic distribution in the latent space. This approach generates high quality samples but cannot map the appointed data samples to the latent space. 

In \cite{chen2019homomorphic, shamsolmoali2022gen, gao2021complementary}, adversarial networks were proposed to generate interpolated auxiliary samples in the latent space. However, these models generally fail to create a convex latent representation while dealing with curved manifolds. To overcome this issue, in \cite{chen2018metrics}, the authors use the magnification factor \cite{bishop1997magnification} to assess the local distortion of the latent space. This factor helps the generated geodesics to seek regions that have dense data distribution on the latent space. In addition, \cite{chen2019fast} introduced a method to detect the shortest distance in a finite graph \cite{li2021adaptive}, similar to the geodesics in the data manifold. This approach is integrated with the latent space through a binary tree data structure \cite{hauser2018principles}. Unlike previous approaches that define geodesics as the shortest path on a graph connecting data points \cite{chen2020learning}, our model works on the Riemannian manifold, and considers the Riemannian metric.\\
\vspace{-7pt}

\noindent{\bf Transformers and Variants.}
Transformer is a attention-based model that can be obtained via transforming CNN filters or manipulating feature maps \cite{jaderberg2015spatial, cohen2016group, schwobel2022probabilistic}. These approaches support limited groups of transformation because they can not accurately compute the cardinality ratio of the transformation. 
\noindent Group equivariant CNNs (\cite{cohen2016group, finzi2020generalizing, romero2020attentive}) are new CNN architectures proposed to address some particular transformations of the input such as rotation. However, they are not able to generalize to more complex compositions of continuous transformation. Because of this, these models cannot be used for more complex transformations in various dimensions.

Our approach stems from the design of STN \cite{jaderberg2015spatial}. A ST layer consists of sub-network prediction parameters with a deformation function. STNs address the problems of geometric variations for a wide range of applications such as filter learning \cite{dai2017deformable} with the following three components:
\begin{itemize}
\item[1.] Localisation Network: This CCN receives the input feature map and produces the transformation parameters $\nu$ that are applied to the feature map via a number of hidden layers.
\item[2.] Parameterised Sampling Grid: The projected transformation parameters are used to generate a sampling grid, from which the input map should be sampled in order to produce the converted output.
\item[3.] Sampler: To perform transformation, the sampler receives the feature map and the sampling grid as inputs and creates an output map by using bilinear interpolation.
\end{itemize}
\noindent In STNs \cite{jaderberg2015spatial}, the input image coordinates are represented with $x\in R^2$, and use parameter $\nu$ to estimate the coordinate transformation $\mathcal{T}_{\nu}(x)$ as
 
\begin{equation}
\mathcal{T}_{\nu}(x)=A_{\nu}
\begin{pmatrix}
x  \\
y  \\
1 
\end{pmatrix}=
\begin{bmatrix}
\nu_{11} & \nu_{12} & \nu_{13} \\
\nu_{21} & \nu_{22} & \nu_{23} 
\end{bmatrix}
\begin{bmatrix}
x  \\
y  \\
1 
\end{bmatrix}.
\label{eq:1}
\end{equation}
in which $A$ is a matrix that use parameter $\nu$ for affine transformation. In STN, the kernel influences the whole image and creates gradient values of all the pixels. This process involves back-propagation across the pixels in the input image and to all the pixels in the transformed one. 

While STN performs well for supervised learning, it has not been widely adopted for generative models. As shown in \cite{lin2018st}, a GAN is used with a STN to compose a foreground object (like a chair), into a background, which gives the image natural appearance. In \cite{eslami2016attend, wan2021high}, VAEs were integrated with STNs for object rendering and image reconstruction to improve diversity, however, these models did not consider disentangled representations, while \cite{skafte2019explicit}, learns disentangled representation by use of perspective inference. In \cite{fang2021transformer}, to more accurately learn the distribution of real data, Transformer-based latent variable model is adopted in CVAE for text generation.
To the best of our knowledge, this is the first attempt to use Riemannian manifold and geodesics learning in generative models with an encoded ST layer for internal representation of latent variables and obtain global context modelling. This approach expectedly leads to more accurate mapping from the input domain to the reconstructed output domain which can be used for a range of computer vision tasks.

\section{Preliminaries on the Riemannian Metric}
We initialize the generated curve by a parameter $s \in [0, 1]$, and assume $\mathscr{M}$ is a distinct Riemannian manifold. $\vartheta:(-\mathcal{E}, \mathcal{E})\rightarrow \mathscr{M}$ is a distinct curve in $\mathscr{M}$. For example, if $\vartheta(0)=r\in \mathscr{M}$, the curve's tangent vector $\vartheta$ at $s=0$ is a function $\acute \vartheta(0)$ defined on the basis of another function $f$ on $\mathscr{M}$ as $\acute \vartheta(0)f=\frac{f\circ \vartheta}{s}\vert_{s=0}$. Different types of tangent vectors on $\mathscr{M}$ and at $r$ are demonstrated via tangent space $T_r \mathscr{M}$. By setting $(I, x)$ as a parametrization of $\mathscr{M}$ at $r$, in addition, for $t\in I$, suppose $x^{(-1)}(t)=(z^1(t),z^2(t),...,z^n(t))$ is the $t$'s local coordinate. Following this, the tangent vector of the coordinate curve $\frac{\partial}{\partial z^i}\vert_r$  at $r$ is determined as  $\frac{\partial}{\partial z^i}\vert_r (f)=\frac{\partial(f\circ x)}{\partial z^i}\vert_r$. We extend a tangent space $T_r \mathscr{M}$ at $r$ by $\{\frac{\partial}{\partial z^i}\vert_r, 1\leq i\leq n\}$. If tangent vectors are mutually orthonormal, thus, the orthonormal basis of tangent space $T_r\mathscr{M}$ is presented by $\{\frac{\partial}{\partial z^i}\vert_r, 1\leq i\leq n\}$.\\
\vspace{-5pt}

Suppose $\mathscr{M}^m$ and $\mathscr{N}^n$ are two distinct manifolds. The differential mapping $\mathscr{O}=\mathscr{M}\rightarrow \mathscr{N}$ at $r$ can be defined under linear mapping $d\mathscr{O}_r:T_r\mathscr{M}\rightarrow T_{\mathscr{O}(r)}$  based on $d\mathscr{O}_r(v)(f)=v(f\circ \mathscr{O})$, as far as $f$ is located on the interpolation curve.
Indeed, a Riemannian manifold is a smooth manifold with a designated metric which assigns to each point $r$ on the tangent space $T_r\mathscr{M}$. Moreover, if $d\mathscr{O}_r:T_r\mathscr{M}\rightarrow T_{\mathscr{O}(r)}\mathscr{N}$ applies at each point $r$, then the mapping $\mathscr{O}$ is isometric immersion, if and only if, $\langle u,v\rangle_r=\langle df_r(u),df_r(v)\rangle_{f(r)}$; $\forall r\in \mathscr{M}$ and $\forall(u,v)\in T_r\mathscr{M}$. 
In Riemannian, a geodesic distance is a parametrized curve $\Upsilon:V\rightarrow \mathscr{M}$ given $\frac{D}{dt}(\frac{d\Upsilon}{dt})$ if $s\in V$ and the length of the shortest path curve can be measured by integrating the norm of its velocity vector $Len=\int_{0}^{1}\Vert \frac{d\Upsilon}{dt}\Vert dt$. In this theory, $\frac{D}{dt}$ is the covariant derivative of a vector field throughout $\Upsilon$ and $\frac{d\Upsilon}{dt}$ denotes the tangent vector along $s$ \cite{hauser2018principles}. To be explicit, this strategy alters the way we estimate distances while maintaining the data structure. Fig. \ref{fig:1} shows an illustrated example that obtained by the Euclidean metric.
\section{Geodesic Manifold Learning}\label{lab:4}
VAEs \cite{higgins2016beta} and some VAE-GAN combinations \cite{ makhzani2015adversarial} can create high-quality samples from latent distributions on a high-dimensional data manifold. On curved surfaces, however, these methods are unable to obtain convex embedding. In \cite{chen2019homomorphic}, the authors proposed separating interpolations from real samples in order to construct interpolations using the distribution of the real data. However, on the low-dimensional manifolds, they create incorrect samples. This issue might be caused by the inadequate capacity of GANs and VAEs in identifying the correlation between the data structures in the latent space. Indeed, GAN based discriminators can only detect the distribution similarity between the generated and real samples, while autoencoders are able to encode manifold curvature information in the latent representations. Thus, the latent embeddings might lead to non-convex representations.

\begin{figure}[t]
\includegraphics[width= 0.88\columnwidth]{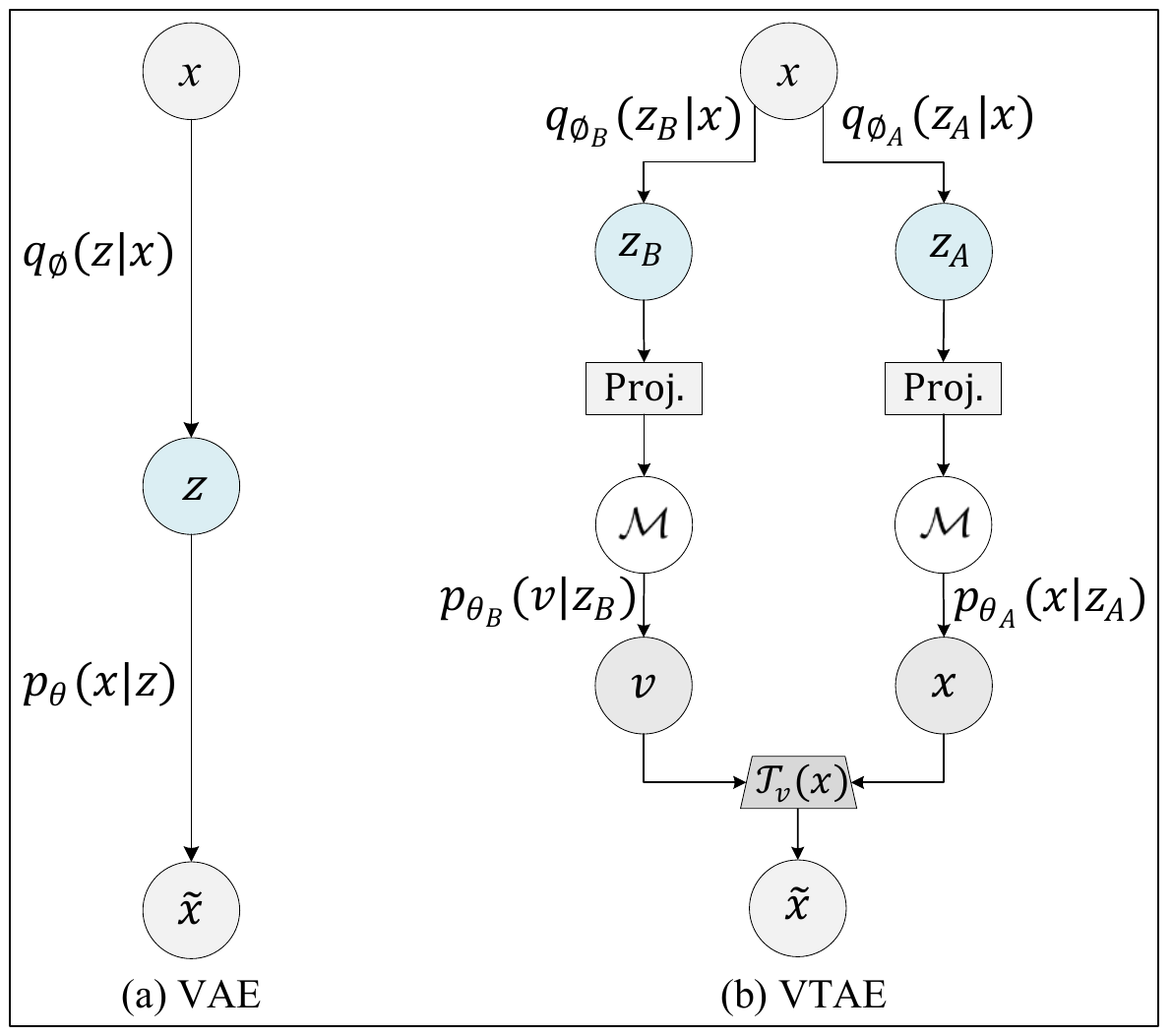} 
\centering
\caption{Conventional VAE architectures and our VTAE. In VTAE, the encoder generates latent spaces $z_A$ and $z_B$ that are projected onto the Riemannian manifold.
${\nu}$ is a transformation parameter in ST-layer $\mathcal{T}_{\nu}$.}   
\label{architecture}
\end{figure}

In VAEs, the data is created by taking a sample from $z$ and subsequently a sample $x$ from the Decoder. To build a flexible model that can deal with the distribution of complex data, $\mu_p$ and $\sigma^2_p$ are designed as deep neural networks. The marginal likelihood is therefore intractable, requiring the use of a variational estimation $q$ to $p(x\vert z)$ as
\begin{equation}
q(z\vert x)=\mathcal{N}(z\vert \mu_q (x),\sigma^2_q(x))
\label{eq:12a}
\end{equation}
in which $\mu_q(x)$ and $\sigma^2_q(x)$ represent different neural networks, as seen in Fig. \ref{architecture} (a).
In the latent space, the univariate Gaussian probability distribution $\mathcal{N}(\mu_q,\sigma^2_q)$ is a node, and the edges are the distances between different distributions. To train VAEs, both Encoder and generative module $p_\theta(x\vert z)p_\theta(z)$ sequentially are trained \cite{eslami2016attend}, and we intend to maximize the ELBO for the best solution:
\begin{equation}
\begin{aligned}
\log p(x)\ge {\mathbb E}_{q_\varnothing(z\vert x)} \Big[\log\frac{p_\theta(x,z)}{q_\varnothing(z\vert x)}\Big]=\\
{\mathbb E}_{q_\varnothing(z\vert x)}\big[\log p_\theta(x\vert z)\big]-KL(q_\varnothing (z\vert x)\Vert p_\theta(z)).
\label{eq:13a}
\end{aligned}
\end{equation}
\noindent The first term in Eq. (\ref{eq:13a}) represents the reconstruction loss between conditional $p_\theta(x\vert z)$ and sample $x$, while the next term represents the KL-divergence between the $q_\varnothing(z\vert x)$ and the previous $p(z)$.
In our model, the base generative module is changed to use two latent variables as a source of information.
\begin{equation}
p(x)=\iint p(x\vert z_A,z_B)p(z_A)p(z_B)dz_A dz_B,
\label{eq:14a}
\end{equation}
in which both the latent variables ($z_A$ and $z_B$) have normal Gaussian priors. We use the following three-step approach with a ST layer for producing new data $x$ (see Fig. \ref{architecture} (b)):
\begin{itemize}
\item[I.]	Obtain samples ($z_A$ and $z_B$) from the latent distribution.
\item[II.]	Decode the obtained samples and use geodesic learning on the Riemannian manifold to acquire $\tilde{x}\sim p(x\vert z_A)$ and $\nu\sim p(x\vert z_B)$.
\item[III.] Apply a ST layer to transform $\tilde{x}$ using parameter $\nu$: ${x}={\mathcal T}_\nu \tilde{x}$.
\end{itemize}

To perform the above steps, we adopt variational inference and latent space $z$ is divided into two smaller latent spaces ($z_A$ and $z_B$), which are represented in ELBO, where the KL-term is also subdivided into two terms.
In this approach, we approximate each latent set of components ($z_A$, $z_B$) separately as $(p(x\vert z_A)\approx q_A (z_A\vert x)$ and $p(x\vert z_B)\approx q_B (z_B\vert x))$.
Our model adopts a conventional manifold learning method \cite {zhang2003nonlinear} which supplies some geometric information for encoding latent representation of data, and a normalizer \cite{chen2020learning} is added to flatten the latent space for better learning outcomes, where $z_A$ is unconditioned on $z_B$. Therefore, the new VAE's loss based on the geodesic distance $GD$ can be written as
\begin{equation}
\begin{aligned}
\ell_{VAE}=& \mathbb{E}_{(z_A, z_B) \sim q_\varnothing(z_A, z_B\vert x})[\log{(p_\theta(x \vert z_A, z_B))}] ~- \\ & {KL}(q_\varnothing(z_A, z_B \vert x)\Vert p_\theta(z_A, z_B)) ~+ \\&\frac{1}{\mathcal{X}^2}\sum_{i,j}(\Vert E(x_i)-E(x_j)\Vert_2 - GD_{ij})^2,
\label{eq:1aa}
\end{aligned}
\end{equation}
in which we train encoder $E$ and decoder $D$ to minimize $\ell_{VAE}$. The input $x$ is embedded to the data manifold $\mathcal{M}$ distribution and a random sample $z_A + z_B=\mu + \sigma \epsilon$ is taken from the data distribution, where $\epsilon \sim \mathcal{N}(0,0.1)$ represents gaussian noise. The Riemannian metric on manifold $\mathcal{M}$ is obtainable via the canonical metric at the Euclidean space $\mathbb{R}^N$ to ensure an isometric immersion. Therefore, to acquire a geodesic distance, the Riemannian geometry is applied on $\mathbb{R}^N$ with the isometric immersion.
$\mathcal{X}$ is the number of input $x$ and $GD_{ij}$ denotes the estimated geodesic distance between two points $(E(x_i)$ and $E(x_j))$. To obtain $GD_{ij}$, we use k-nearest neighbors at each point and to determine the shortest distances between two points we adopt the conventional manifold learning \cite{zhang2003nonlinear}. Therefore, $E$ with $GD_{ij}$ is trained to attain a convex latent space representation while $D$ has to learn the missing curvature information on the distribution of manifolds.

\subsection{Manifold Interpolation Network}
To decode significant objects, and provide progressive transformations, our model estimates geodesics by interpolating in the latent space and decode these interpolations into the data space. The easiest interpolation method is linear interpolation as ($(1-s)\times z_1+s\times z_2$), however, it is not appropriate for most geodesic learning cases. To effectively configure the geodesic curves, a fixed-point iteration scheme and a bounded quadratic function is used. Inspired by \cite{kalatzis2020variational}, we apply cubic functions to bundle interpolation on the diversity of the latent space, $g(s)=as^3 + bs^2 + cs + d$, where $a, b, c$ and $d$ are four $J$-dimensional vectors and $J$ denotes the latent coordinates' dimension for producing a curve. More specific, our model trains a geodesic curve $g(s)$ that joints two points $z_1$ and $z_2$, therefore the model must be confined to meet $g(0)=z_1$ and $g(1)=z_2$. As a result, our interpolation curve is smooth, and it passes over high-probability areas in the latent space. 

\subsection{Insertion Loss}
Based on the geodesic assumption \cite{lesniak2018distribution}, the parameter $s$ is equivalent to the curve' s length $\Upsilon(s)$ and we can generate auxiliary samples over a curve on data manifold $\mathcal{M}$ by decoding the interpolation network $g(s)$.
Assume $\mathcal{M}\subset \mathbb{R}^N$ as a Riemannian differentiable manifold. Therefore, the decoder of the interpolation network induces Riemannian metric in the latent space $\mathbb{R}^N$. If the shortest path curve $\Upsilon: V\rightarrow \mathcal{M}, [x_1(s),x_2(s),...,x_N(s)$ is the rectangular coordinate system of $\Upsilon_{(s)}$ on $\mathbb{R}^N$, thus the length of tangent vector $\sqrt{\sum_i \frac{d_{x_i}(s)^2}{d_t}}$ is constant $\forall s\in V$. The decoder takes the geodesic (interpolation) curve of high density regions as input $G(s)=D(g(s))$, and the insertion loss is written as
\begin{equation}
\ell_{ins}=\frac{1}{n}\sum_{i=1}^n\left(\frac{\Vert \acute G(s_i)\Vert_2}{\bar{s}(\Vert \acute G(s)\Vert_2)}-1\right)^2,
\label{eq:2aa}
\end{equation}
in which $n$ is the number of the points on the manifold $\mathcal{M}$ and $\bar{s}(\Vert \acute G(s)\Vert_2)=\frac{1}{n}\sum_{i=1}^n\Vert \acute G(s_i)\Vert_2$.  $\acute G(s)$ represents the derivative of the decoder output $G(s)$ concerning $\bar s(mean_s)$ and $s$. Therefore, the insertion loss makes $G(s)$ to interpolate uniformly between two data samples.

\subsection{Geodesic Distance Loss}\label{lab:4c}
Given that $G(s)$ has a uniform interpolation, we must guarantee $G(s)$ is the shortest curve. To seek the shortest curve connecting two points, the following theorem is used.
Assume $\Upsilon:V\rightarrow \mathcal{M}$ is a parametrized curve on a Riemannian manifold, $h(\rlap{U}-)$ is a local coordinate space with $\Upsilon_s\subset h(\rlap{U}-)$.$[z_1,z_2,...,z_m]$ denotes the local coordinate of $h(\rlap{U}-)$, and the Cartesian coordinate system of $\Upsilon_s$ in $\mathbb{R}^N$ is represented by $[x_1(s),x_2(s),...,x_N(s)]$. Therefore, the shortest curve from $z_1$ to $z_2$ on a $\mathcal{M}$ is computed via the following function:
\begin{equation}
\Upsilon_s^{shortest}=\underset{\Upsilon_s}{argmin}\int_{0}^{1}\sqrt{(\Upsilon_{s}^{'},\mathcal{M}_{\Upsilon_s}\Upsilon_s^{'})d_s},
\label{eq:3aa}
\end{equation}
in which $\Upsilon(0)=z_1,\Upsilon(1)=z_2$, and $\Upsilon_s:[0,1]\rightarrow\mathcal{M}$.
$\Upsilon_s^{'}=\frac{\partial \Upsilon_s}{\partial s}$ is the first derivative of the curve, where $\mathcal{M}_\Upsilon$ denotes the Riemannian metric. We can estimate the shortest distances by optimizing the curve energy \cite{hauser2018principles} as
 \begin{equation}
\Upsilon_s^{shortest}=\underset{\Upsilon_s}{argmin}\int_{0}^{1}(\Upsilon_s^{'}, \mathcal{M}_{\Upsilon_s}\Upsilon_s^{'})d_s,
\label{eq:4aa}
\end{equation}
The inner product can be expressed as
 \begin{equation}
\begin{aligned}
L(\Upsilon_s,\Upsilon_s^{'}, \mathcal{M}{\Upsilon_s})&=\sum_{i=0}^d\sum_{j=1}^d{\Upsilon^{'}_s}^{(i)}{\Upsilon^{'}_s}^{(j)}\mathcal{M}_{\Upsilon_s}^{(ij)}\\
&=(\Upsilon_s\otimes \Upsilon_s^{'})^{\top}vec[\mathcal{M}_{\Upsilon_s}]
\label{eq:5aa}
\end{aligned}
\end{equation}
in which $\otimes$ denotes the standard Kronecker product, $\top$ represents the projection from $\mathbb{R}^N$ to the tangent space $T_r\mathscr{M}$, and $vec[.]$ stacks the matrix's columns into a vector. Based on the Lagrange equation $L$, we can minimize the absolute curve energy as follows:
\begin{equation}
\underbrace {~~ \frac{\partial L}{\partial \Upsilon_s}~~ } =\underbrace { ~~ \frac{\partial}{\partial s}\frac{\partial L}{\partial \Upsilon_s^{'}} ~~ }
\label{eq:6aa}
\end{equation}
It can be written as
 \begin{equation}
\begin{aligned}
\frac{\partial}{\partial s}\frac{\partial L}{\partial \Upsilon_s^{'}}=\frac{\partial}{\partial s}(2.\mathcal{M}_{\Upsilon_s}\Upsilon_s^{'})=2\left[\frac{\partial \mathcal{M}_{\Upsilon_s}}{\partial s}\Upsilon_s^{'}+\mathcal{M}_{\Upsilon_s}\Upsilon_s^{''}\right].
\label{eq:7aa}
\end{aligned}
\end{equation}
The left and right hand sides of Eq. \ref{eq:6aa} are respectively equal to
\begin{equation}
\begin{aligned}
\frac{\partial L}{\partial\Upsilon_s}&=\underbrace{\frac{\partial}{\partial\Upsilon_s}((\Upsilon_s^{'}\otimes \Upsilon_s^{'})^\top vec[\mathcal{M}_{\Upsilon_s}])}\\
\frac{\partial}{\partial s}\frac{\partial L}{\partial \Upsilon_s^{'}}&=\underbrace{2\left[{\Upsilon_s^{'}}^\top \frac{\partial vec[\mathcal{M}_{\Upsilon_s}]}{\partial \Upsilon_s} \Upsilon_s^{'} +\mathcal{M}_{\Upsilon_s}\Upsilon_s^{''}\right]}.
\label{eq:8aa}
\end{aligned}
\end{equation}
On the bases of Eq. \ref{eq:6aa} and Eq. \ref{eq:8aa}, a curve $\Upsilon_s$ connecting two points is geodesic if its second derivative $\Upsilon_s^{''}$ is orthogonal to the tangent vector and can be written as

\begin{equation}
\begin{aligned}
\Upsilon_s^{''}&=-\frac{1}{2}\mathcal{M}_{\Upsilon_s}^{-1} \bigg[2.{\Upsilon_s^{'}}^\top\frac{\partial vec[\mathcal{M}_{\Upsilon_s}]}{\partial \Upsilon_s}{\Upsilon_s^{'}}\\ & ~~~~\; - \frac{\partial vec[\mathcal{M}_{\Upsilon_s}]^\top}{\partial \Upsilon_s}(\Upsilon_s^{'}\otimes \Upsilon_s^{'})\bigg].
\label{eq:9aa}
\end{aligned}
\end{equation}
In our model, the encoder links a data point to its local coordinates and we can write the geodesic distance loss as
\begin{equation}
\begin{aligned}
\ell_{geo}=\frac{1}{n}\sum_{i=1}^{n}\Vert G^{''}(s_i).D^{'}(g(s_i))\Vert_2,
\label{eq:10aa}
\end{aligned}
\end{equation}
in which $D$ denotes the decoder operation and $\acute D(g(s_i))$ is the decoder $D$ derivative at $g(s_i)$. $G^{''}(s_i)$ represents a multi-dimensional vector on the basis of $[x_1^{''}(s),x_2^{''}(s),...,x_N^{''}(s)]$ and $\acute D(g(s))$ denotes an $M\times N$ matrix consistent with $[\frac{\partial h}{\partial z_1}\vert\Upsilon_s,\frac{\partial h}{\partial z_2}\vert\Upsilon_s,...,\frac{\partial h}{\partial z_m}\vert\Upsilon_s]$. Our proposed model utilizes the shortest length between two points to assure $G(s)$ is a minimizing geodesic. We estimate the length of the curve using the velocity addition at $s_i$. To find the shortest length curve, the minimized geodesic distance loss is computed as
\begin{equation}
\ell_{mgeo}=\sum_{i=1}^{n}\Vert G^{'}(s_i)\Vert_2.
\label{eq:11aa}
\end{equation}

\noindent To decrease the computation cost, the following approximation is used
\begin{equation}
\begin{aligned}
&G^{'}(s) \approx \frac{G(s+\bigtriangleup s)+G(s-\bigtriangleup s)}{2\bigtriangleup s}, \\
&G^{''}(s) \approx \frac{G(s+\bigtriangleup s)+G(s-\bigtriangleup s)-2G(s)}{\bigtriangleup s^2}
\label{eq:12aa}
\end{aligned}
\end{equation}

\noindent To compute  $D^{'}(g(s))$ , we utilize the Jacobian of $D$. Based on the above losses, we can write the total loss of the proposed interpolation network as
\begin{equation}
\begin{aligned}
\ell_{total}=w_1 \ell_{ins} + w_2 \ell_{geo} + w_3 \ell_{mgeo},
\label{eq:13aa}
\end{aligned}
\end{equation}

\noindent in which $w_1=1$, $w_2=0.01$, and $w_3=10$ denote the weights for regularizing the three losses with default values to have robust performance. 

\section{Transformer layer}
In our model, for each pixel, the interpolation network generates auxiliary sample locations next to the given pixel. Then, geodesic sampling \cite{gong2012geodesic} is used to determine these locations. It has been practically tested that sampling on the geodesic between the source and target samples yielded higher consistency in view of geometrical features. Consequently, such samples can be seen as latent representations.
Therefore, based on the transformation parameter $\mathcal{T}_\nu$ of a sample point $x$, the local linear approximation can be computed.
Since standard convolution only applies a local neighborhood at a time, it is incapable of capturing global contextual information. Thus, in the spatial dimensions, we use non-local filtering \cite{wang2018non} in ST layer.
Moreover, to ensure invertibility, we parametrize the transformation. One rotation angle of $\alpha$, one shear parameter of $\rho$, two scaling factors of $\mathfrak{s}_x$ and $\mathfrak{s}_y$, and two translation parameters of $\tau_x$ and $\tau_y$ are all included as
\begin{equation}
\mathcal{T}_{\nu}(x)=
\begin{bmatrix}
\cos(\alpha)& -\sin(\alpha)  \\
\sin(\alpha)& \cos(\alpha)   \\
\end{bmatrix}
\begin{bmatrix}
1 & \rho \\
0 & 1 
\end{bmatrix}
\begin{bmatrix}
\mathfrak{s}_x & 0  \\
0 & \mathfrak{s}_y
\end{bmatrix}
\begin{bmatrix}
\tau_x \\
\tau_y
\end{bmatrix},
\label{eq:15aa}
\end{equation}
in which the scale factors must both be positive. The matrix exponential is another simple and more elegant technique. In other words, we parameterize the transformation's velocity rather than its transformation in Eq. \ref{eq:1}
\begin{equation}
\mathcal{T}_{\nu}(x)=exp\Bigg(
\begin{bmatrix}
\nu_{11} & \nu_{12} & \nu_{13} \\
\nu_{21} & \nu_{22} & \nu_{23}
\end{bmatrix}\Bigg)
\begin{bmatrix}
x  \\
y  \\
1 
\end{bmatrix}\Bigg .
\label{eq:16aa}
\end{equation}

Often, we have no idea about which transformation classes are appropriate for data disentanglement, therefore, use of a highly adaptable class of transformations is an obvious option.
In addition, to solve the nonconvex problem of transformers, generally the Tikhonov method is used as a regularizer \cite{jiang2019linearized}. However, the performance of Tikhonov is sensitive to the choice of the learning parameters. In our ST layer, projected gradient descent is used to solve this problem, whereas both the gradient and the projections rely on the basis of the entropy regularizer (ER) \cite {benamou2015iterative}. By replacing the Tikhonov with the entropy regularizer, we use the weight of the regularizer which is a significant hyperparameter to improve the convergence.

\begin{figure}[t]
\centering
\includegraphics[width=1\columnwidth]{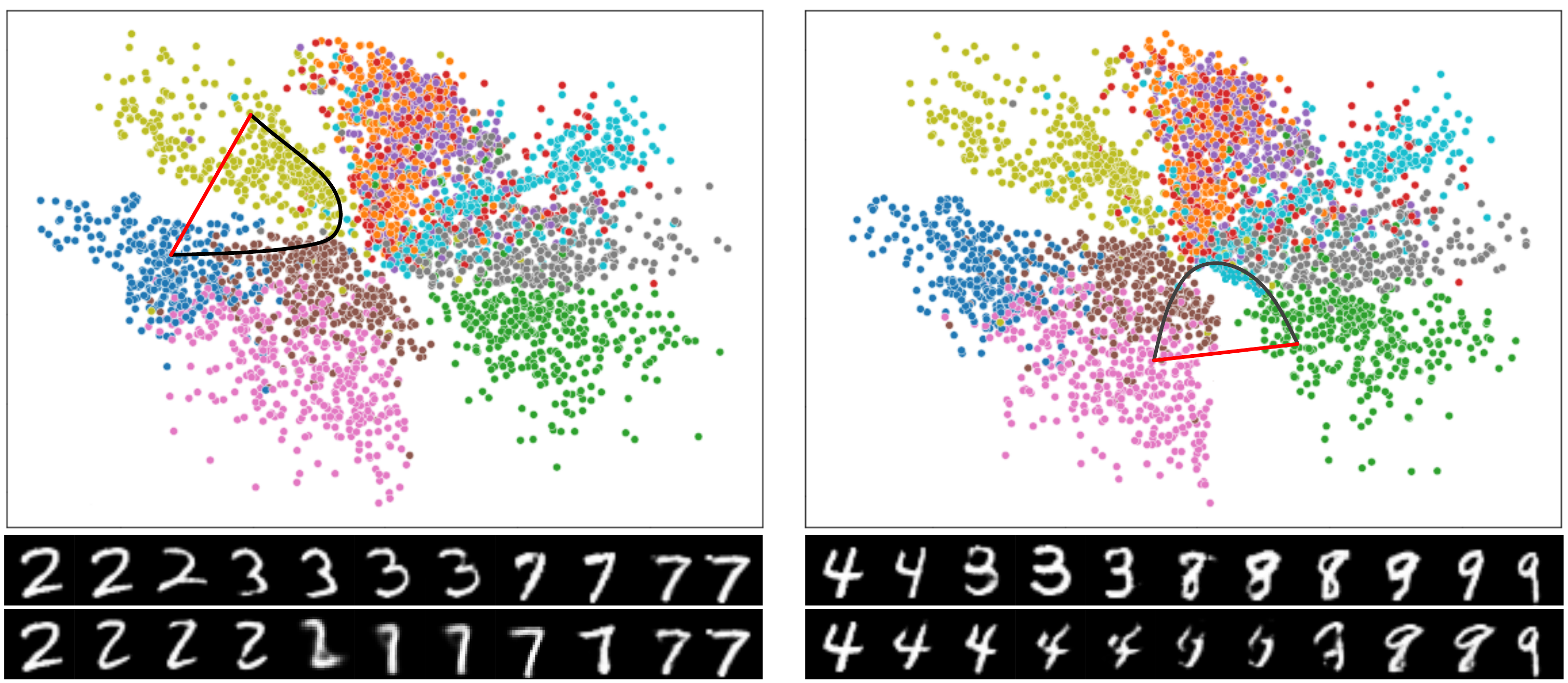} 
\caption{Image decoding by linear/geodesic interpolation: Top - Interpolations plotted in the latent space. Black lines show geodesics and red lines represent linear interpolation. Middle - Decoded images with geodesic interpolation. Bottom - Decoded images with linear interpolation.}   
\label{fig:MNIST LS}
\end{figure}

\section{Experiments}
In this section, the performance of our VTAE is evaluated on image interpolations, image reconstructions, and compared against those of several state-of-the-art models. The models are evaluated on three grayscale image datasets - MNIST \cite{lecun1998gradient}, FashionMNIST \cite{xiao2017fashion}, EMNIST \cite{cohen2017emnist}, and three natural image datasets - CelebA \cite{liu2018large}, Places2 \cite{zhou2017places}, and CIFAR-10 \cite{krizhevsky2009learning}. For all the datasets, we use the standard training-testing splits. \\
\vspace{-7pt}

In our experiments we use a VAE with the following architecture: the encoder contains six layers. Four convolution layers (convolution sizes = $[32, 64, 128, 256]$, strides = [2, 2, 2, 2], and kernel sizes = [4, 4, 4, 4]) plus two linear reshape layers. ReLU activation function is used in all the layers except the last two layers. The decoder contains five layers (convolution sizes = [256, 128, 64, 32,1], strides = [2, 2, 2, 2, 1], and kernel sizes = [4, 4, 4, 4, 4]). ReLU activation function is used in the first four layers, whereas the sigmoid activation functions is used in the final layer, and batch normalization is used on all the layers except the last one. 
To train our network, Adam with default values ($\beta$1 = 0.9, $\beta$2 = 0.999) is used to optimize the weights, with learning rate of 0.0001, weight decay = 0.001 and momentum = 0.9.
Our VTAE contains four ST layers that are trained together with the encoder and the decoder, and the loss is determined as the difference between the decoder's output and the target.
The weights of the KL and appearance matching losses can be derived from the ELBO.
In order to obtain better hyper-parameters, grid search is used to determine the training configuration that performs best on each validation set.

We use the same neural network for the VAE architecture throughout the experiments. Additionally, all the models use one stochastic layer with 50 latent variables. 
We train our interpolation module for 20,000 iterations on the grayscale image datasets. On the other hand, for the natural image datasets, our interpolation network is trained for 35,000 iterations.
We implement the proposed VTAE in Pytorch \footnote{The code is publicly available at: https://github.com/pshams55/VTAE, after the paper has been accepted for publication.}, and run it on a RTX 3090 GPU.
\begin{figure*}[t]
\centering
\includegraphics[width=2.05 \columnwidth]{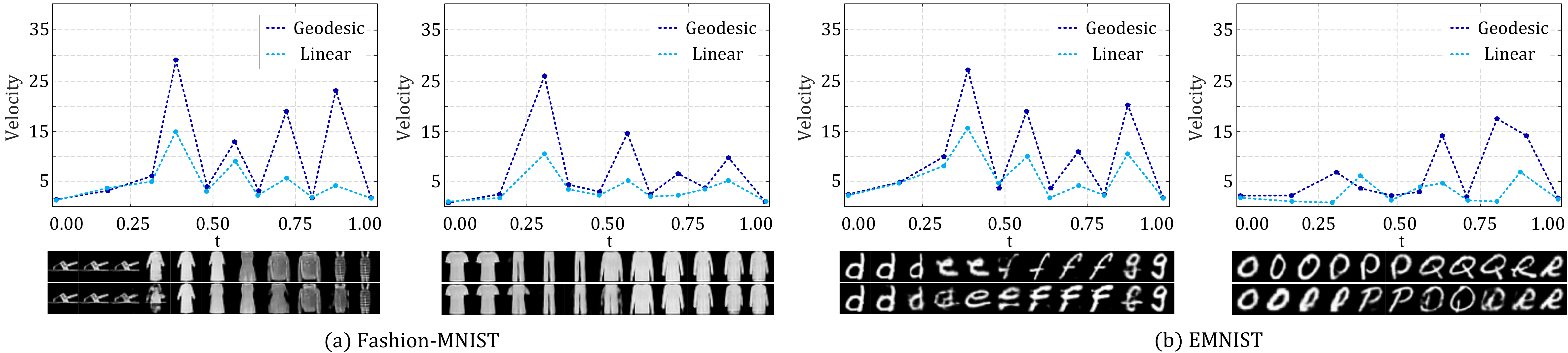} 
\caption{Decoded images over the Fashion-MNIST and EMNIST data samples. Top: the corresponding velocity curves of the interpolation. Middle: Decoded images with geodesic interpolation. Bottom: Decoded images with linear interpolation.}   
\label{fig:EF-MNIST-LS}
\end{figure*}
\begin{figure}
\centering
\includegraphics[width=1\columnwidth]{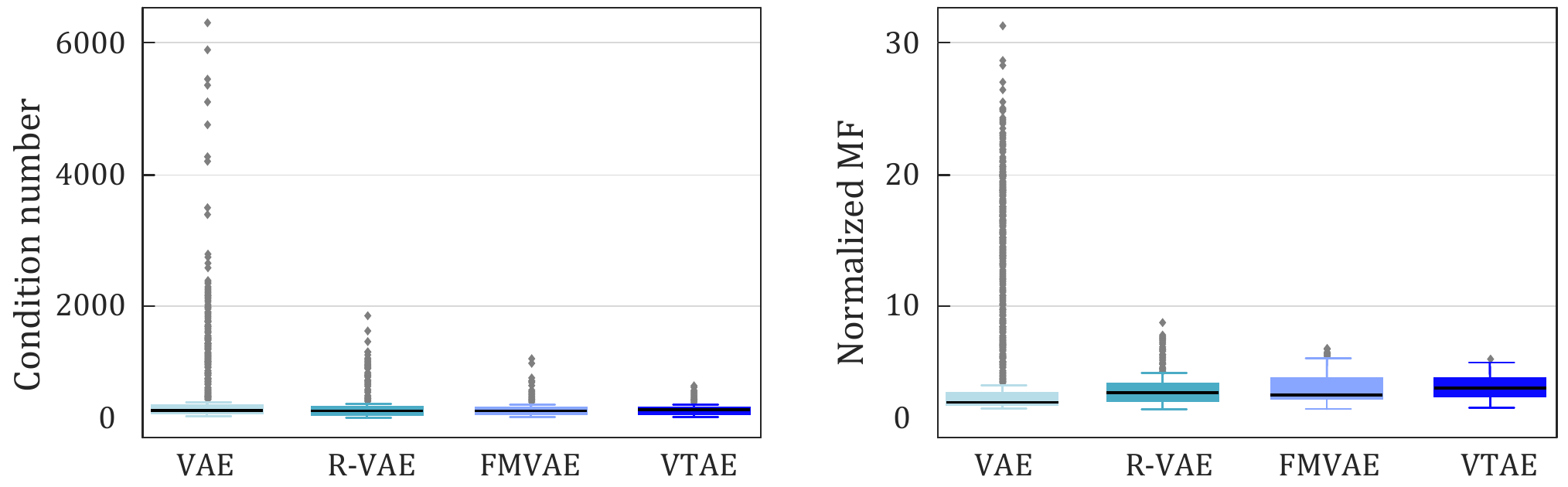} 
\caption{For various encoders, condition number and normalized MF scores were obtained. The lower condition number and normalised values are better. The box plots were created using 8,000 randomly generated samples.}   
\label{fig:MIG}
\end{figure}

\begin{figure*}
\centering
\includegraphics[width=1.97\columnwidth]{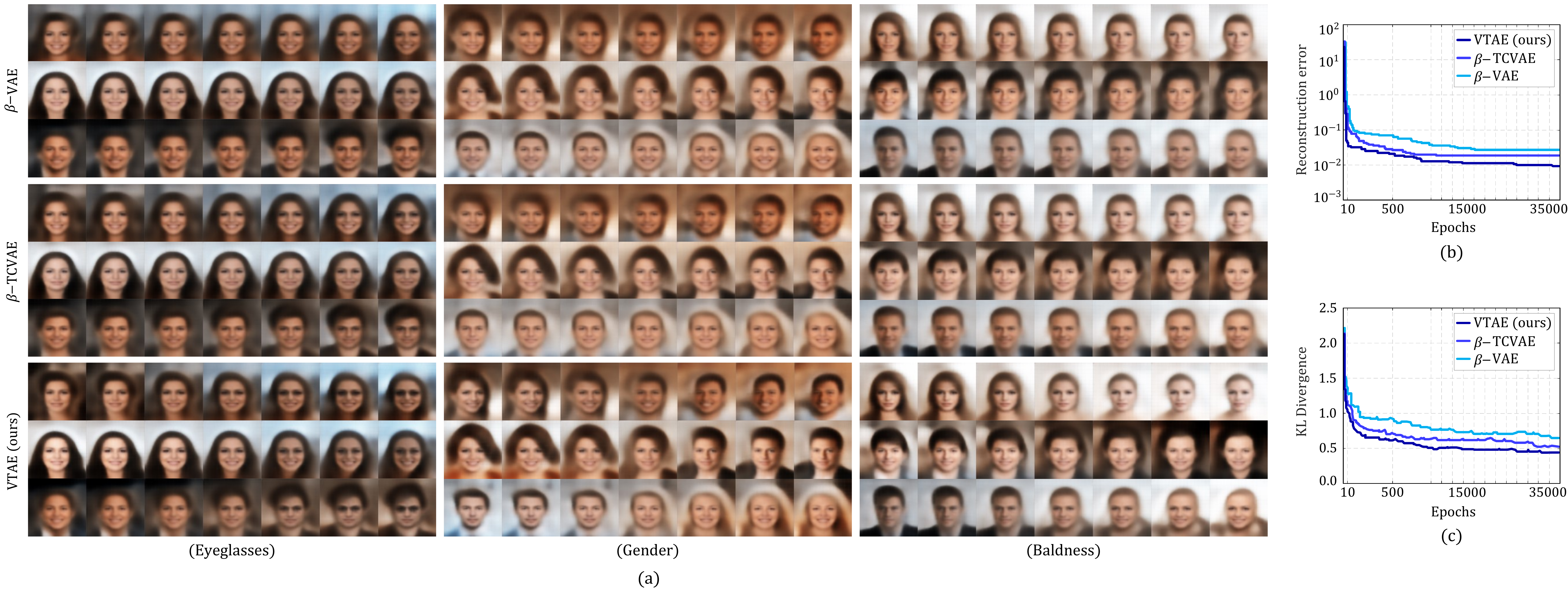}
\caption{Qualitative comparisons On CelebA. (a) The interpolated images for three VAE models are shown: our proposed VTAE, $\beta$-TCVAE, and $\beta$-VAE ($\beta$=250). Since attributes can only be manifested with a latent variable, we demonstrate a one-sided traversal. Each row contains a seed image from which the latent values were inferred. As the results show, VTAE produces sharper samples and better reconstruction quality in comparison to other models.
(b) and (c) are the reconstraction error and KL Divergence curves during the training.}   
\label{fig:interpolate-celeb-1}
\end{figure*}

\begin{figure*}[t]
\centering
\includegraphics[width=14.94cm, scale=0.6]{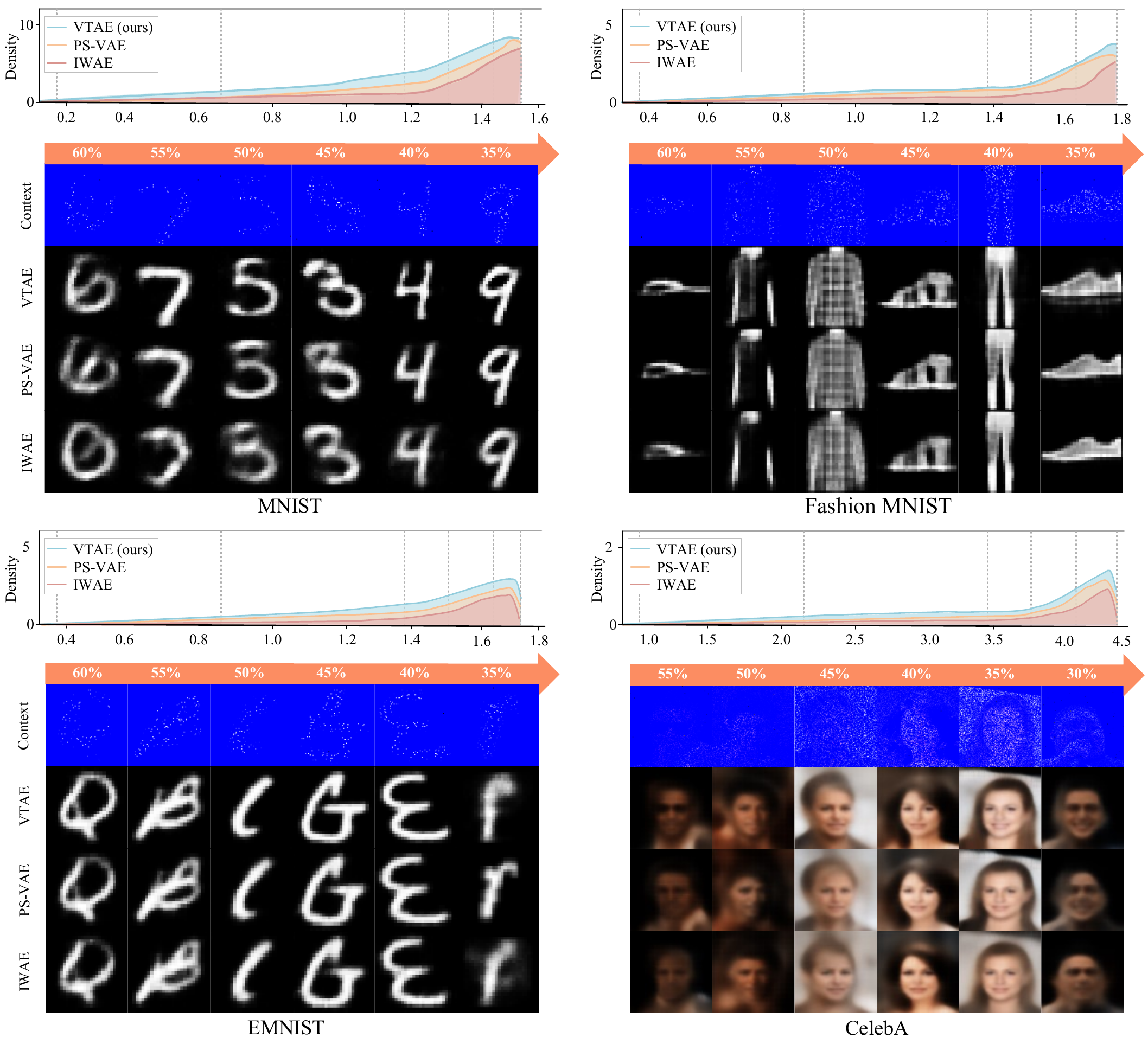} 
\caption{Log-likelihood and qualitative results on 4 datasets with different noise levels. The top row represents the log-likelihood distribution for VTAE, PS-VAE, and IWAE. The following images stand for the percentages of noise (first row), VTAE target predictions (second row), PS-VAE target predictions (third row), and IWAE target predictions.}   
\label{fig:5}
\end{figure*}
\begin{figure*}[!h]
\centering
\includegraphics[width=1.63\columnwidth]{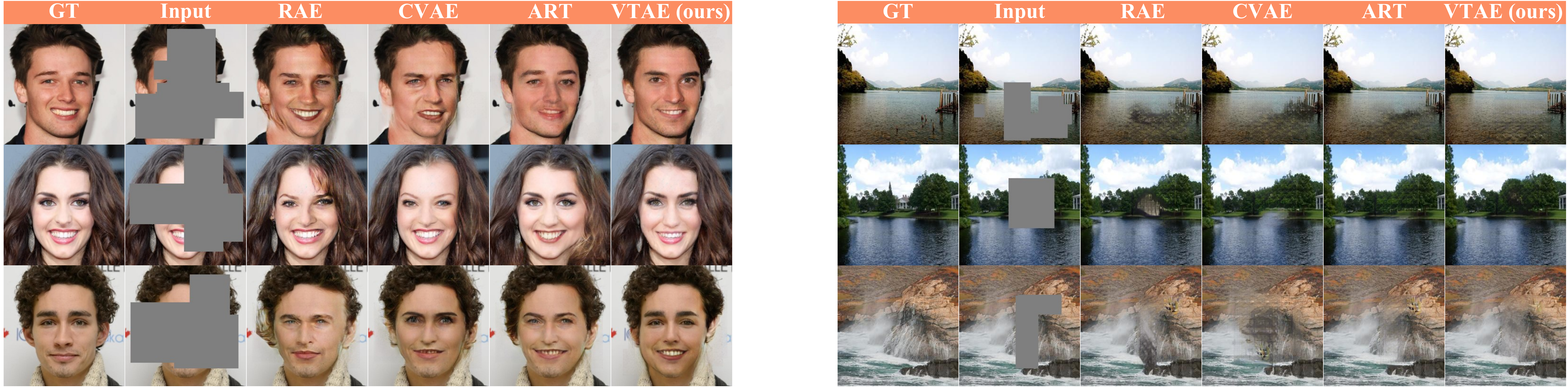} 
\caption{Qualitative results and comparisons with different baselines on CelebA ($256\times256$) (left) and Places2 (right) test sets. Our method outperforms other approaches in terms of preserving both structures and textures.}   
\label{fig:img-completion}
\end{figure*}


\subsection{Image Interpolation}
To illustrate the efficacy of our approach in image interpolation, we choose two images at random as the start and end points in the data manifold for each dataset \cite{shao2018riemannian}. To achieve image interpolation between these two images, we use different interpolation algorithms. For the MNIST dataset, In Fig. \ref{fig:MNIST LS}, the generated linear and geodesic interpolant trajectories are shown as a 2-D donut distribution. We find that although our geodesic method discovers a longer route in the latent space than the standard linear interpolation, the intermediate spots appear more realistic and give a semantically relevant morphing between its ends because it does not traverse a region where there is no data. With our interpolation procedure, we assure that the created curves follow the data in all directions. As seen in Fig. \ref{fig:MNIST LS}, even while the linear and geodesic trajectories are both on the data manifold in some cases, the interpolation results are different.

We use the same strategy to interpolate images on the Fashion-MNIST and EMNIST datasets, as shown in Fig. \ref{fig:EF-MNIST-LS}. The smoothness of different interpolation algorithms is assessed using the velocity metric discussed in Section (\ref{lab:4}). When moving across classes, our model can always provide identifiable images, but linear interpolation causes ghosting with two objects. This explains why our geodesic learning method effectively interpolates on the geodesic curve.

The Riemannian metric (which is based on the decoder's Jacobian, as discussed in Section (\ref{lab:4c})) is used to estimate local angles, length, and surface area. It also has many inherent features of a manifold. In order to further evaluate the models, condition numbers and magnification factors (MFs) are computed using the Riemannian metric \cite{chen2020learning}. The condition number, which represents the ratio of the longest to shortest axes and the sensitivity of the likelihood functions is represented by MF. MF is a scaling coefficient for projecting from the latent space (Riemannian) to the observation space (Euclidean). Because MFs of various methods cannot be compared directly, MFs are normalized by their means. Therefore, a model is highly invariant to the Riemannian metric if the conditional number and the normalized MF approach 1. More specifically, the normalized MF and the conditional number are metrics for determining if the metric is constant and close to 1. To visualise the condition number and MF, similar to \cite{chen2020learning}, we adopt the hierarchical prior in VAEs proposed in \cite{klushyn2019learning}. 
Using the normalised MF and condition number metrics, we compare the performance of our proposed model to that of the other baselines on the MNIST dataset in Fig. \ref{fig:MIG}. VTAE, unlike the other models, learns a latent space in which Euclidean distances are close to geodesic distances.

\begin{table*}
\renewcommand{\arraystretch}{1.1} 
\centering
\caption{\footnotesize \uppercase{Comparison against the advanced log-likelihood-based generative models. The performance is given in bits per dimension (better off lower).}} \label{tab:1}
\small{
\begin{tabular}{l|cccccc}     \hline\hline
\multirow{ 3}{*}{Method} & \multicolumn{6}{c}{Log-Likelihood}  \\   \cline{2-7}   
            & MNIST & FashionMNIST & EMNIST & CIFAR-10  & CelebA &Places2 \\ 
& $28\times28$ &  $28\times28$&  $28\times28$ &  $32\times32$ &  $64\times64$ & $256\times256$ \\    \hline
VAE \cite{kingma2013auto}& 0.91 & 1.75& 1.94 & 3.21  & 2.28  & 0.97 \\   
$\beta$-VAE \cite{higgins2016beta}&  0.77  & 1.62 & 1.73 & 3.06 & 2.17  & 0.80  \\  
IWAE \cite{ im2017denoising} &  0.84  & 1.68 &1.85 & 3.14 & 2.23 & 0.88  \\  
PS-VAE \cite{yang2021learning}&  0.67  & 1.41& 1.69 & \bf{2.88} & 2.09  & 0.75  \\   
$\beta$-TCVAE \cite{chen2018isolating}& 0.72  & 1.53 & \bf{1.61} & 2.97 & 2.11 & 0.73 \\                                                  
VTAE (ours) &  \bf{0.65}& \bf{1.36 }& 1.64 & 2.90  & \bf{2.02}  & \bf {0.66}  \\   \hline
\end{tabular}
}
 \end{table*}
\subsection{Image Reconstruction}
To evaluate the performance of our model for sample interpolation and reconstruction, we have no metric to assess system performance as we do not fully know the underlying data manifold. We can only provide qualitative evaluations on the reconstruction and interpolation quality. From Fig. \ref{fig:interpolate-celeb-1} (a), independent of the latent dimensions, $\beta$-VAE, $\beta$-TCVAE, and VTAE have smooth interpolations, however, VTAE generates sharper human faces due to its interpolation capability, compared against the other models. VTAE also detects a number of additional factors missed by $\beta$-VAE, such as colour and image saturation. Moreover, VTAE learns a smooth and continuous transformation across a broader range of factors than $\beta$-VAE and $\beta$-TCVAE. In some images, $\beta$-TCVAE reaches the same level of reconstruction quality as ours, but in other examples, as can be seen in the $1^{st}$ to $2^{nd}$ columns in Fig. \ref{fig:interpolate-celeb-1} (a), it struggles in accurate interpolation and reconstruction.
The training curves for CelebA are shown in Fig. \ref{fig:interpolate-celeb-1} (b) and (c). As a result, the reconstruction error and KL divergence of our model are lower than those of the other baselines and Our VTAE model appears to have the best log-likelihood. Furthermore, the latent variables derived from our detection network during training more closely resemble the genuine distribution utilized during testing. This is demonstrated by the KL divergence in Fig. \ref{fig:interpolate-celeb-1} (c) during the training of the recognition network of various VAE-based models. We find that the KL divergence of the recognition network trained with our model is smaller, therefore, it minimizes the latent variable mismatch between the training and testing pipelines.\\
\vspace{-7pt}

The negative log-likelihood of the test data is widely used to assess the performance of deep generative models, and the negative log-likelihood is normalized by the number of dimensions for interpretability. By normalizing, we can estimate bits per dimension or get a lower bound on the predicted bits required for lossless compression using a Huffman code $p(x)$.
In our experiments, we adopt the importance-weighted stochastic variational inference to approximate the log-likelihood, in which higher compression rates result from better distribution estimates. Table \ref{tab:1} reports the likelihoods for our model and the prior models on six datasets. Results indicate that VTAE has better performance in comparison with the other baselines. In particular, for the CelebA and Places2 datasets, our model outperforms $\beta$-TCVAE and PS-VAE.

To assess the performance of our model for image reconstruction (denoising) \cite{li2022matrix}, we train our model to generate clean inputs with noise injected at the input level to reconstruct a distorted input image. When the input is distorted, the conventional VAE requires marginalizing the conditional distribution of the encoder, making the training criteria intractable. Instead, similar to \cite{im2017denoising}, we set up a modified training criterion that, in the case of corruption, leads to a tractable bound.
In this experiment, for grayscale datasets, each pixel of an image is sampled from (0, 1) according to its pixel intensity value, and from the CelebA we normalized the images such that each pixel value ranges between [0, 1]. We compare the performance of VTAE, IWAE \cite{ im2017denoising}, and PS-VAE \cite{yang2021learning} with corruption distributions of varying noise levels.
The batch size is set to 100, and the results is reported based on log-likelihood.
We use a standard technique to select a noise distribution. To corrupt the original images, we add salt and pepper noise to the greyscale datasets and Gaussian noise to the CelebA.

In Tables \ref{tab:3} and \ref{tab:4}, we respectively show the average negative log-likelihood for various levels of corruption on the MNIST and CelebA datasets. As we can see, VTAE outperforms the other models in most experiments. Moreover, it seems that the models are insensitive to the noise types: Gaussian or salt and pepper. Indeed, the models are more sensitive to noise intensity instead of noise type. 

\begin{table}[h!]
\renewcommand{\arraystretch}{1.1} 
\centering
\caption{\footnotesize \uppercase{Average negative log-likelihood with various noise levels on MNIST (better off the lower).}} \label{tab:3}
\begin{tabular}{l|ccc}     \hline\hline
\multirow{ 2}{*}{Method} & \multicolumn{3}{c}{Noise Level}  \\   \cline{2-4}   
            & 10$\%$ & 25$\%$& 40$\%$   \\    \hline
VAE \cite{kingma2013auto}& 0.91 & 1.02& 1.13   \\    
IWAE \cite{ im2017denoising} &  0.84 & 0.90 & 0.98  \\ 
PS-VAE \cite{yang2021learning}&  0.69 & 0.75& 0.83    \\                                                      
VTAE (ours) &  \bf{0.65}& \bf{0.68}& \bf{0.72}   \\   \hline
\end{tabular}
 \end{table}
\begin{table}[h!]
\renewcommand{\arraystretch}{1.1} 
\centering
\caption{\footnotesize \uppercase{Average negative log-likelihood with various noise levels on CelebA.}} \label{tab:4}
\begin{tabular}{l|ccc}     \hline\hline
\multirow{ 2}{*}{Method} & \multicolumn{3}{c}{Noise Level}  \\   \cline{2-4}   
            & 10$\%$ & 25$\%$& 40$\%$  \\    \hline
VAE \cite{kingma2013auto}& 2.37 & 2.48& 2.61   \\     
IWAE \cite{ im2017denoising} &  2.31 & 2.39 & 2.50   \\ 
PS-VAE \cite{yang2021learning}&  2.09 & 2.16& 2.23    \\                                                      
VTAE (ours) &  \bf{2.04}& \bf{2.08}& \bf{2.14}   \\   \hline
\end{tabular}
\end{table}
\begin{table}[h!]
\centering
\caption{ \footnotesize \uppercase{Quantitative evaluation on Places2 dataset.}}
\begin{tabular}{l|ccc}     \hline\hline
Method & PSNR $\uparrow$ & NIQE $\downarrow$& IS $\uparrow$ \\ \hline
RAE \cite{ghosh2019variational} & 19.61 & 6.49&23.84   \\ 
GICA \cite{yu2018generative} & 18.96  & 6.34&23.95  \\
IWAE \cite{ im2017denoising}& 20.04 & 6.12&24.56   \\ 
CVAE \cite{sohn2015learning} & 21.45  & 5.81&25.17  \\ 
ART \cite{wan2021high} & 22.41  & 4.95& \bf{26.13}   \\   
VTAE (ours) & \bf{22.78 } & \bf{4.72 }&26.06  \\   \hline
\end{tabular}
\label{tab:2}
\end{table}

This is understandable given that when the noise level is high, the model loses the information needed to reconstruct the original input, and there will be a significant difference between the distributions of the (corrupted) training and test datasets. Fig. \ref{fig:5} reveals a visual representation of the image log-likelihoods with a different noise level, along with the distribution results of VTAE, IWAE, and PS-VAE. 
Results show that VTAE provides more accurate reconstruction compared to the other two baselines.
Even when the target pixels are filled, VTAE compare to PS-VAE generates smoother reconstruction which is due to use of geodesic learning. 

To reconstruct an image with large holes, we train the network to reconstruct the central region of $128\times128$ in a $256\times256$ image for regular holes (as used in \cite{yu2018generative}), with pixel-wise reconstruction loss and VAE loss as the objective function. Based on the masked image, the generative path infers the latent distribution of the holes. Therefore, the reconstruction loss targets the visible regions (in this experiment, we use a network with seven convolution layers for encoder and eight convolution layers for decoder with two fully connected layers).
Fig. \ref{fig:img-completion} shows examples with different scales' holes to evaluate the performance of our model on CelebA and Places2. As the results show, CVAE \cite{sohn2015learning} and ART \cite{wan2021high} generate acceptable results for most cases as the conditional prior is highly focused on the maximum latent probability. However, our model generates more plausible results by sampling from the two latent spaces. Our model uses the surrounding textures, resulting in more realistic reconstructed images with fewer artifacts than the baselines. Because of the ST layers, our model can recognize contextual image structures, and use information from surrounding areas to promote the generation.\\
\vspace{-7pt}

\noindent{\bf Quantitative comparisons:} We compare our model with other methods in Table \ref{tab:2}. The comparison is based on test images from the Places2, with a $128\times128$ mask in the center. To evaluate the low-level differences between the reconstructed output and the ground truth, the Peak Signal-to-Noise Ratio (PSNR), Natural Image Quality Evaluator (NIQE) \cite{mittal2012making}, and Inception Score (IS) \cite{salimans2016improved} are used.
It can be seen that our model outperforms other baselines. We outperforms ART, particularly in the PSNR and NIQE metrics. In comparison to other approaches, our model can eliminate general noise artifacts while preserving the input details.\\
\vspace{-7pt}

\begin{figure}[t]
\centering
\includegraphics[width=0.98\columnwidth]{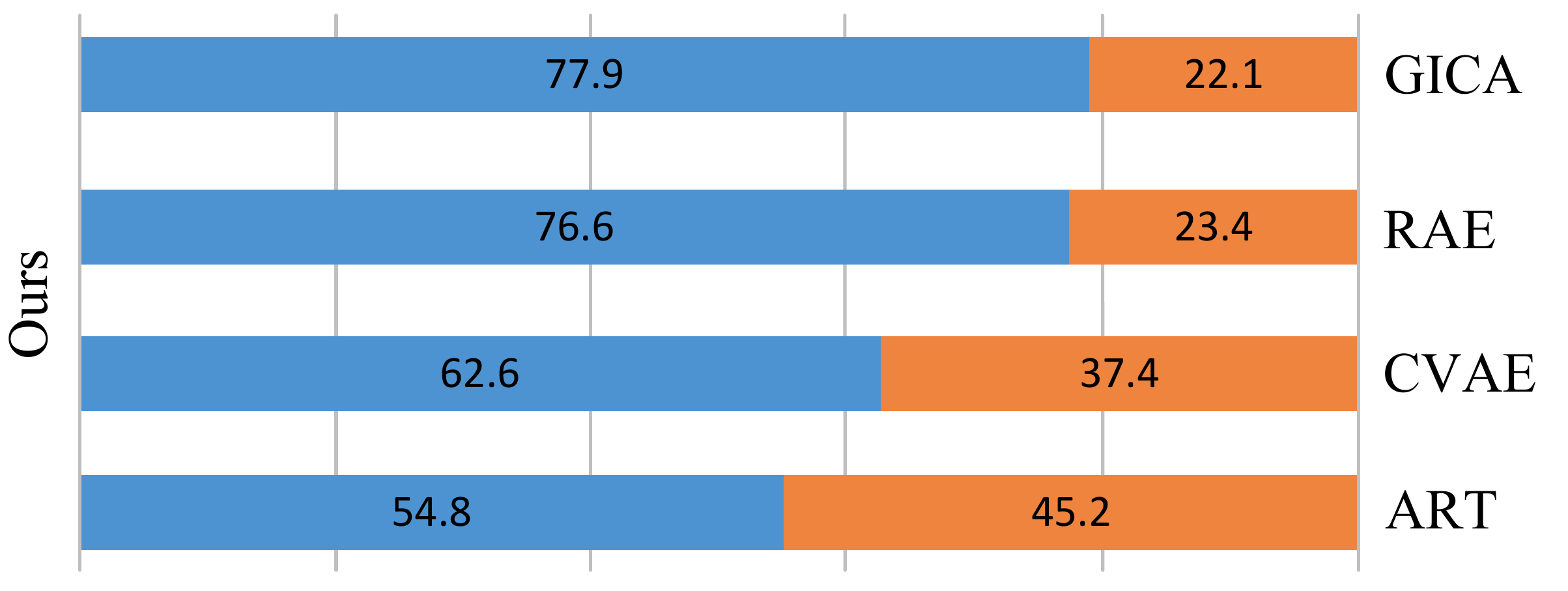} 
\caption{The results of a study on human perception. The values represent the preference for the comparison pair.}   
\label{fig:user-study}
\vspace{10pt}
\end{figure}

\noindent {\bf Human perceptual study:} We also perform a user study against other models to more accurately assess the subjective quality. We take 20 masked images at random from the Places2 test set. Different approaches are used to produce reconstructed outputs for a test image. In this assessment, participants are given two separate reconstructed images at the same time, one created by our model and the other by one of the baselines. Participants are asked to select the most photorealistic and visually natural image. We collect responses from 23 participants and calculate the ratios of each approach using the data given in Fig. \ref{fig:user-study}. Our method is of 67.98$\%$ likelihood to be selected.

\begin{figure}
\centering
\includegraphics[width=0.98\columnwidth]{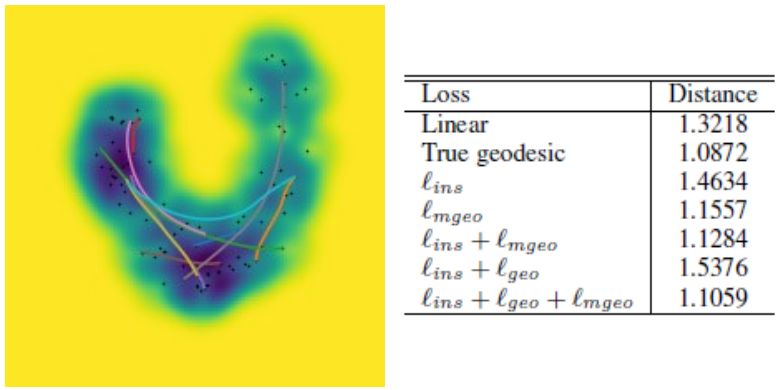} 
\caption{Distance measurement in the latent space for images of ’0’ in MNIST and the interpolation curve with various losses by our model.}   
\label{fig:distance}
\end{figure}

\subsection{Ablation Study}
On the MNIST dataset, we conduct ablation studies to understand the impact of various losses proposed in our methodology, such as the insertion loss, geodesic distance loss, and minimizing geodesic distance loss. From digit '0', we take 100 samples and compute pair-wise distances between the samples, shown in Fig. \ref{fig:distance}. The geodesic distance indicates a solid clustering structure in this example.
We find that the insertion loss encourages uniform interpolation. In comparison to linear interpolation, our network constructs a shorter path that is close to a true geodesic without causing the geodesic distance loss. Despite the fact that our network can quickly converge to a reasonably decent output, the geodesic accuracy is not satisfactory. Our interpolated points are fine-tuned to move when combined with the geodesic distance loss. Our finding is supported by the anticipated curves given in Fig. \ref{fig:distance}. The generated curve with all three losses produces the best results, and the predicted curve length is the most similar to the length of a true geodesic.\\
\vspace{-9pt}

\begin{figure}[h!]
\centering
\includegraphics[width=0.98\columnwidth]{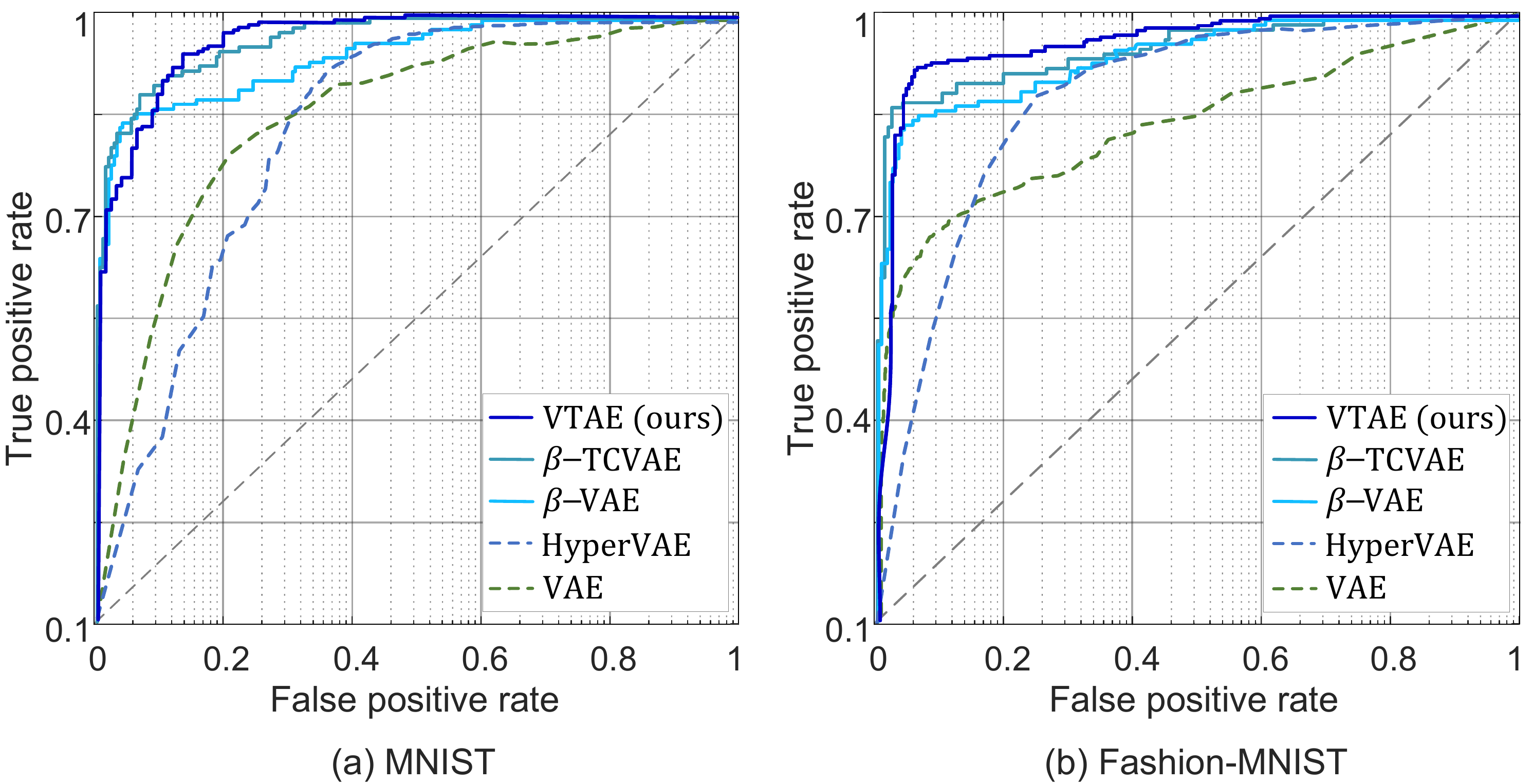} 
\caption{ROC curves show the performance of several encoders on two datasets.}   
\label{fig:ROC}
\vspace{-10pt}
\end{figure}

As an additional measure of VTAE's performance, we use the mean squared error between the reconstructed and the original images as a threshold to explore outliers. To compute ROC curves, the true and false-positive rates of the resulting labels are compared to the ground truth. Fig. \ref{fig:ROC} compares the ROC curves of several VAE models on the MNIST and Fashion-MNIST datasets. Based on the findings, our VTAE method manages to converge to a higher ROC and outperform $\beta$-VAE and $\beta$-TCVAE on image reconstruction. Specifically, VTAE outperforms $\beta$-VAE by $12\%$ on the MNIST dataset and $9\%$ on the Fashion-MNIST dataset.\\
\vspace{-9pt}

\begin{figure}[h!]
\centering
\includegraphics[width=0.99\columnwidth]{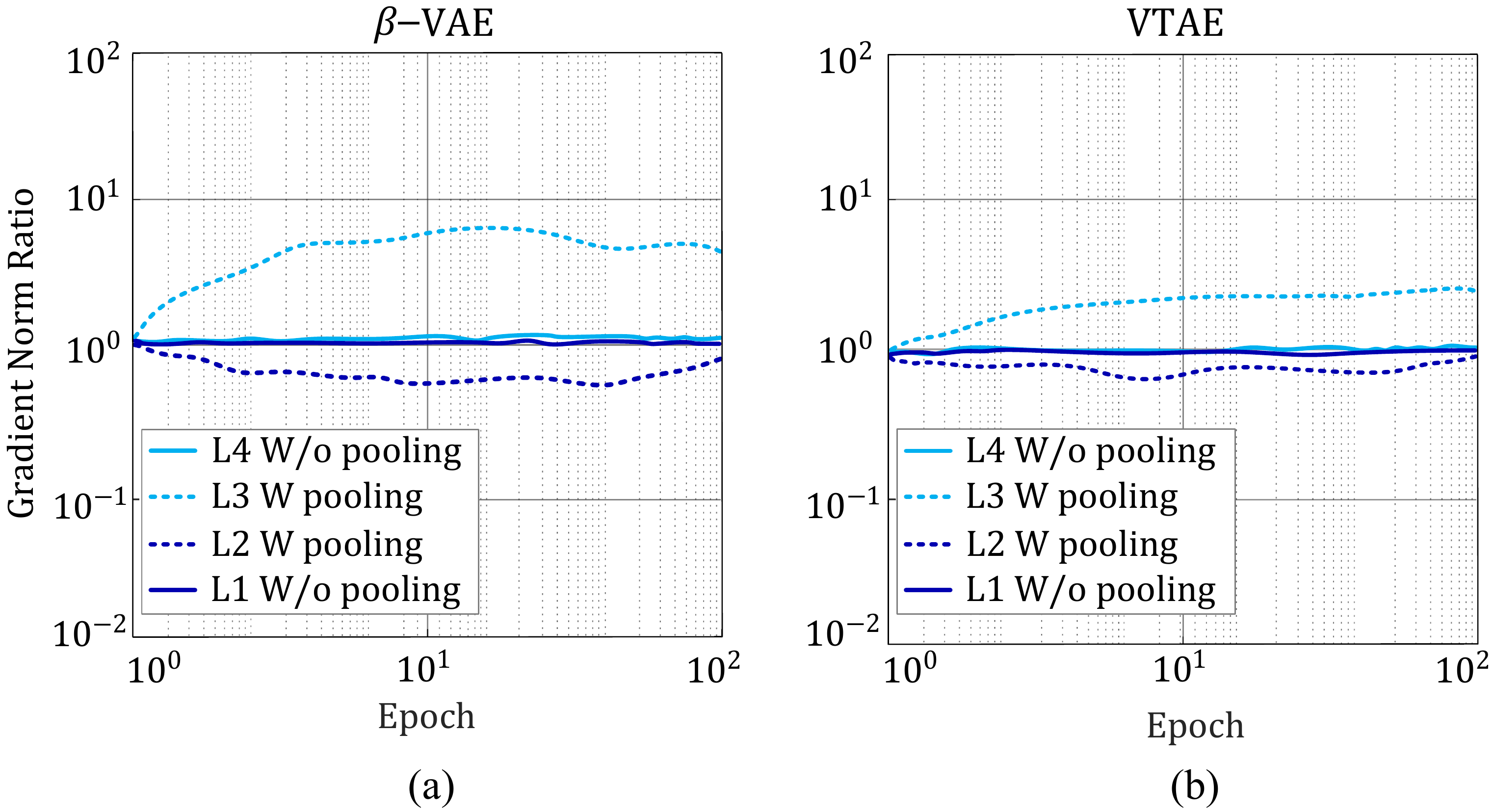} 
\caption{Training evaluation of $\beta$-VAE in comparison with VTAE. It shows the ratios of gradient norms for first 100 epochs. The solid lines denote the without pooling layers, where the dashed lines are the layers with pooling.}   
\label{fig:7}
\end{figure}

In Fig. \ref{fig:7}, the ratios of $\beta$-VAE and VTAE are evaluated over the training epochs. In this experiment, we use a basic VAE architecture. The encoder contains 4 layers, 2 with pooling layers of changing dimensions and the other 2 layers without pooling. We train the networks for 100 epochs without any learning rate decay. This investigation shows the importance of the entropy regularizer and initialization for the construction of a norm-preserving network. 
Despite the fact that $\beta$-VAE are norm-preserving at the early stage, the broad range covered by the gradient becomes higher and moves away from 1, while the parameters are updated. The results show that VTAE has a stronger norm-preserving ability.
In our proposed model, the encoded ST layer, and geodesic learning on a Riemannian manifold, are helpful to improve the performance of our proposed model, otherwise the network fails to impose the manifold invariance or cannot be robust against the off-manifold noise. The ST layer with entropy regularizer is helpful to improve VTAE performance, otherwise, the network is less robust to distribution shifts. Moreover, our proposed geodesic interpolation network effectively decodes meaningful objects, provides a progressive transformation, and reflects the underlying structure of the dataset.


\section{Conclusion} 
In this paper, we designed a new architecture for variational autoencoder with an encoded ST layer that uses consistent interpolations with constrained geodesic learning to enhance representation learning capability of our model. To have a smooth interpolation on the latent space, we utilize the extracted geometric knowledge to regularize our network and proposed an interpolation network with the insertion loss and the geodesic distance loss. Experiments on a number of datasets demonstrate that our model achieves high reconstruction quality while also creating a meaningful, and compact latent representation.

\footnotesize{
\bibliographystyle{IEEEtran}
\bibliography{IEEEabrv,IEEEexample}}

\end{document}